\documentclass{article}

\usepackage{PRIMEarxiv}

\usepackage[utf8]{inputenc} 
\usepackage[T1]{fontenc}    
\usepackage{hyperref}       
\usepackage{url}            
\usepackage{booktabs}       
\usepackage{amsfonts}       
\usepackage{nicefrac}       
\usepackage{microtype}      
\usepackage{xcolor}         
\usepackage{amsmath}
\usepackage{graphicx,color}
\usepackage{bbm, dsfont}
\usepackage{float}
\usepackage{cases}
\usepackage{algorithm}
\usepackage[noend]{algpseudocode}
\usepackage{longtable}
\usepackage{tabularx}
\newcommand{\RR}{\ensuremath{\mathbb{R}}}
\makeatletter
\def\BState{\State\hskip-\ALG@thistlm}

\usepackage{cite}
\usepackage{amssymb,amsfonts}
\usepackage{textcomp}
\usepackage{xcolor}
\def\BibTeX{{\rm B\kern-.05em{\sc i\kern-.025em b}\kern-.08em
    T\kern-.1667em\lower.7ex\hbox{E}\kern-.125emX}}

\usepackage{multirow}
\graphicspath{{media/}}     

\title{Distributed MCMC inference for Bayesian Non-Parametric Latent Block Model}

\author{
Reda Khoufache \\
        {\small \textit{DAVID Lab}}\\ 
        {\small \textit{Paris-Saclay University, UVSQ}}\\
        {\small Versailles, France}\\
        {\small reda.khoufache@uvsq.fr}
\and
\textbf{Anisse Belhadj} \\
        {\small \textit{DAVID Lab}}\\ 
        {\small \textit{Paris-Saclay University, UVSQ}}\\
        {\small Versailles, France}\\
        {\small med-anisse.belhadj@outlook.com \vspace{0.2cm} }
\and
\textbf{Hanene Azzag} \\
        {\small \textit{LIPN (CNRS UMR 7030)}}\\ 
        {\small \textit{Sorbonne Paris Nord University}}\\
        {\small Villetaneuse, France}\\
        {\small azzag@univ-paris13.fr}
\and
\textbf{Mustapha Lebbah} \\
        {\small \textit{DAVID Lab}}\\ 
        {\small \textit{Paris-Saclay University, UVSQ}}\\
        {\small Versailles, France}\\
        {\small mustapha.lebbah@uvsq.fr}
}

\begin{document}
\maketitle

\begin{abstract}
In this paper, we introduce a novel Distributed Markov Chain Monte Carlo (MCMC) inference method for the Bayesian Non-Parametric Latent Block Model (DisNPLBM), employing the Master/Worker architecture. Our non-parametric co-clustering algorithm divides observations and features into partitions using latent multivariate Gaussian block distributions. The rows are evenly distributed among workers, which exclusively communicate with the master and not among themselves. DisNPLBM demonstrates its impact on cluster labeling accuracy and execution times through experimental results. Moreover, we present a real-use case applying our approach to co-cluster gene expression data. The code source is publicly available at \url{https://github.com/redakhoufache/Distributed-NPLBM}.
\end{abstract}

\section{Introduction}
Given a data matrix, where rows represent observations and columns represent variables or features, co-clustering, also known as bi-clustering aims to infer a row partition and a column partition simultaneously. The resulting partition is composed of homogeneous blocks.  When a dataset exhibits a dual structure between observations and variables, co-clustering outperforms conventional clustering algorithms which only infers a row partition without considering the relationships between observations and variables. Co-clustering is a powerful data mining tool for two-dimensional data and is widely applied in various fields such as bioinformatics \cite{articlebio}. 

To tackle the co-clustering problem, the Latent Block Model (LBM) was introduced by 
\cite{article}. This probabilistic model assumes the existence of hidden block 
components, such that elements that belong to the same block independently follow 
identical distribution. A Bayesian Non-Parametric extension of the LBM (NPLBM) was 
introduced in \cite{Meeds07nonparametricbayesian}. This model makes two separate 
priors on the proportions and a prior on the block component distribution, which allows to automatically estimate the number of co-clusters during the inference process. In \cite{goffinet2021coclustering}, the authors present a BNP Functional LBM  which extends the recent LBM \cite{BENSLIMEN201897} to a BNP framework to address the multivariate time series co-clustering.

To infer parameters of NPLBM, the Collapsed Gibbs sampler introduced in \cite{10.2307/1390653}, is a Markov Chain Monte Carlo (MCMC) algorithm that iteratively updates the column partition, given the row partition, and vice versa. It samples row and column memberships sequentially based on their respective marginal posterior probabilities. The collapsed Gibbs sampler is an efficient MCMC algorithm because the co-cluster parameters are analytically integrated away during the sampling process.
MCMC methods have the good property of producing asymptotically exact samples from the target density. However, these 
techniques are known to suffer from slow convergence when dealing with large datasets.  

Distributed computing consists of distributing data across multiple computing nodes (workers), which allows parallel computations to be performed independently. Distributed computing offers the advantage of accelerating computations and overcoming memory limitations. Existing programming paradigms for distributed computing, such as Map-Reduce consist of a map and reduce functions. The map function applies the needed transformations on data and produces intermediate key/value pairs. The reduce function merges the map's function results to form the output value. The Map-Reduce job is executed on master/workers architecture, where the master coordinates the job execution, and workers execute the map and reduce tasks in parallel.

This paper proposes a new distributed MCMC-based approach for NPLBM inference when the number of observations is too large. We summarize our contributions: \textbf{(1)} We have developed a new distributed MCMC inference of the NPLBM using the Master/Worker architecture. The rows are evenly distributed among the workers which only communicate with the master. \textbf{(2)} Each worker infers a local row partition given the global column partition. Then, sufficient statistics associated with each local row cluster are sent to the master. \textbf{(3)} At the master, the global row partition is estimated. Then, given the global row partition, the column partition is estimated. This allows the estimation of the global co-clustering structure. \textbf{(4)} Theoretical background and computational details are provided. 
\section{Related Work}
Numerous scalable co-clustering algorithms have been proposed in the literature. The first distributed co-clustering (DisCo) 
using Hadoop is introduced in  \cite{4781146}. In \cite{Folinoinproceedings}, authors devised a parallelized co-clustering 
approach, specifically designed to tackle the high-order co-clustering problem with heterogeneous data. Their methodology 
extends the approach initially proposed in \cite{5342422}, enabling the computation of co-clustering solutions in a parallel 
fashion, leveraging a Map-Reduce infrastructure. In \cite{5576169}, a parallel simultaneous co-clustering and learning (SCOAL) 
approach is introduced, also harnessing the power of Map-Reduce. This work focuses on predictive modeling for bi-modal data. 
In \cite{7145441}, introduces a distributed framework for data co-clustering with sequential updates (Co-ClusterD). The 
authors propose two distinct approaches to parallelize sequential updates for alternate minimization co-clustering algorithms. 
However, it's worth noting that these approaches are parametric and assume knowing a priori the true numbers of row and column 
clusters, respectively, which are unknowable in real-life applications. One of the main challenges in distributing Bayesian 
Non-Parametric co-clustering lies in efficiently handling and discovering new block components.  
\section{Bayesian Non-Parametric Latent Block Model}
\subsection{Model definition}
Let $n$, $p$, and $d$ be positive integers, and let $X = (x_{i,j})_{n,p} \in \mathbb{R}^{n\times p \times d}$ be the observed dataset. Here, $n$ represents the number of rows, $p$ is the number of columns, and $d$ denotes the dimension of the observation space. Let $\mathbf{z} = (z_i)_n$ be the row membership vector (row partition), where each $z_i$ is a latent variable such that $z_i = k$ signifies that the $i$-th row $x_{i,\cdot}$ belongs to the row cluster $k$. Similarly, let $\mathbf{w}=(w_j)_p$ be the column membership vector (column partition), where $w_j = l$ indicates that the $j$-th column $x_{\cdot,j}$ belongs to the 
column cluster $l$. The NPLBM is defined as follows:
$$
\begin{gathered}
    x_{i, j} \mid \{ z_{i}, w_{j}, \theta_{z_{i}, w_{j}}\} \sim F\left(\theta_{z_{i},w_{j}}\right),\\ 
    \theta_{z_{i}, w_{j}} \sim G_{0},\,\,z_{i}\left|\pi \sim \operatorname{Mult}(\pi),\,\, w_{j}\right| \rho \sim \operatorname{Mult}(\rho),\\
    \pi \sim \mathrm{SB}(\alpha),\,\, \rho \sim \mathrm{SB}(\beta). 
\end{gathered}
$$
According to this definition, the observation $x_{i,j}$ is sampled by first generating the row proportions $\pi \sim \mathrm{SB}(\alpha)$ and column 
proportions $\rho \sim \mathrm{SB}(\beta)$ according to the Stick-Breaking ($\mathrm{SB}$) process \cite{10.2307/24305538} parameterized by concentration 
parameters $\alpha>0$ and $\beta>0$ respectively. Secondly, sampling the row and column memberships $\mathbf{z}$ and $\mathbf{w}$ from the Multinomial 
distribution ($\operatorname{Mult}$) parameterized by $\pi$ and $\rho$, respectively. Then, sampling the block component parameter $\theta_{z_i,w_j}$ from the base distribution $G_0$. Finally, drawing the cell value $x_{i,j}$ that belongs to the block $(z_i,w_j)$ from the component distribution $F(\theta_{z_i,w_j})$. We assume that $F$ is the multivariate Gaussian distribution (i.e, $\theta_{k,l} = (\mu_{k,l}, \Sigma_{k,l})$, with $\mu\in \RR^{d}$ and $\Sigma_{k,l}\in \RR^{d\times d}$ a positive semi-definite matrix), and $G_0$ is the Normal Inverse Wishart \cite{murphy2007conjugate} (NIW) conjugate prior with hyper-parameters $(\mu_0, \kappa_0, \Psi_0, \nu_0)$.  
\subsection{Inference}
The goal is to estimate the row and column partitions $\mathbf{z}$ and $\mathbf{w}$ given the dataset $X$, the 
prior $G_0$, and the concentration parameters $\alpha$ and $\beta$, by sampling from the joint posterior 
distribution $\textrm{p} (\mathbf{z},\textbf{w}|X, G_0, \alpha, \beta)$. However, direct sampling from this 
distribution is intractable but can be achieved using the collapsed Gibbs Sampler introduced in \cite{10.2307/1390653}.
Given initial row and column partitions. The inference process consists of alternating between updating 
the row partition given the column partition and then updating the column partition given the row partition. At each iteration, to update the row partition $\mathbf{z}$, each $z_i$ is updated sequentially by sampling from $\textrm{p}(z_i|\mathbf{z}_{-i}, \mathbf{w}, X, G_0, \alpha)$, where 
$\mathbf{z}_{-i} = \{ z_r | r\neq  i\}$. The column partition update is similar to the row partition update. The complete algorithm and computation 
details of the inference process are given in \cite{goffinet:tel-03491716}. 
\section{Proposed inference}
The main objective of our method is to make the inference scalable when the number of observations becomes 
too large. The rows are distributed evenly over the workers. At each iteration, we alternate 
between two levels: 
\subsection{Worker level}\label{sec worker_level}
\begin{sloppypar}
Let $E$ be the number of workers, $n^{e}$ be the number of rows in worker $e$, $X^{e}=(x^e_{i,j})_{n^e\times p}\in \RR^{n^e\times p\times d}$ the local dataset in worker $e$, each cell $x^e_{i,j}$ is a $d$-dimensional vector. Let $\mathbf{z}^{e}=(z_i^e)_{n^e}$ be the local row  partition (i.e. $z_i^{e} = k$
means that the $i$-th row  of $e$-th worker belongs to the $k$-th local row  cluster). At this level, each local row membership $z_i^e$ is updated given other local row  memberships $\mathbf{z}_{-i}^{e} = \{z_r^e| r\neq i\}$ by sampling from $\text{p}(z_i^e|\mathbf{z}_{-
i}^{e}, \mathbf{w},X^e, G_0, \alpha)\propto$:
\begin{subnumcases}{}
   n_{k}^{e}\mathrm{p}\left(x_{i,\cdot}^e \mid\mathbf{w}, \mathbf{x}_{k,.}^{e}, G_{0}\right) & existing row cluster k, \label{eq w1bis}
   \\
   \alpha \mathrm{p}\left(x_{i,\cdot}^e \mid \mathbf{w}, G_{0}\right), & new row cluster, \label{eq w1bis1}
\end{subnumcases}
where $n^{e}_{k}$ is the size of local row cluster $k$ in worker $e$, $x_{i,\cdot}^e$ is the $i$-th row  of 
worker $e$, and $\mathbf{x}_{k,.}^{e} = \{x_{i,\cdot}^e|z_i^e = k\}$ the content of local row cluster $k$
in worker $e$. Since $G_0$ is a prior conjugate to $F$, the joint prior and posterior predictive distributions needed in \ref{eq w1bis} and \ref{eq w1bis1} are computed analytically \cite{goffinet2021coclustering}. After having updated the local row  partition, for a given row  cluster $k$ in worker $e$, for each column $j$, we compute the following sufficient statistics:
    \begin{align}
    T_{k,j}^{e} &= \frac{1}{n_{k}^{e}} \sum_{i=1,z_{i}^{e}=k}^{n^{e}} x_{i,j}^{e} \in \mathbb{R}^d,\label{eq 10bis2}\\
    S_{k,j}^{e} &= \sum_{i=1,z_{i}^{e}=k}^{n^{e}} ( x_{i,j}^{e}- T_{k,j}^{e})(x_{i,j}^{e} - T_{k,j}^{e})^T \in \mathbb{R}^{d\times d},\label{eq 10bis3}
    \end{align}
    
    where $(\cdot)^T$ denotes the transpose operator. We let 
    $\mathcal{S}^e = \left\{ (T_{k,j}^e, S_{k,j}^e)\mid(k,j)\in \{1,\cdots, K^e\}\times\{1,\cdots,p\} \right\},$
    the set of sufficient statistics, where $K^e$ is the number of row  clusters inferred in worker $e$. Finally, the sufficient statistics and sizes of each cluster are sent to the master. The DisNPLBM inference process at the worker level is described in Algorithm \ref{DisNPLBMRow}, which represents the Map function.
\end{sloppypar}
\begin{algorithm}
\caption{DisNPLBM inference at worker level}\label{DisNPLBMRow}
\begin{algorithmic}[1]
\State \textbf{Input}: $X^{e}_{n^{e}\times p \times d}$, $\alpha$, $G_0$, $\mathbf{z}^e$, and $\mathbf{w}$.
\BState \textbf{For} $i\leftarrow  1$ \textbf{to} $n^{e}$ \textbf{do}:
\BState \indent Remove $x^e_{i,.}$ from the its local row cluster.
\BState \indent Sample $z_{i}^{e}$ according to Eq. \ref{eq w1bis} and Eq. \ref{eq w1bis1}.
\BState \indent Add $x^e_{i,.}$ to its new local row cluster.
\BState \textbf{For} $k\leftarrow  1$ \textbf{to} $K^{e}$ \textbf{do}:
\BState \indent \textbf{For} $j\leftarrow  1$ \textbf{to} $p$ \textbf{do}:
\BState \indent \indent Compute $T_{k,j}^e$ and $S_{k,j}^e$ as defined in Eq. \ref{eq 10bis2} and Eq. \ref{eq 10bis3}, respectively.
\State \textbf{Output}: Updated row partition $\mathbf{z}^e$, sufficient statistics $\mathcal{S}^e$, sizes of each cluster.
\end{algorithmic}
\end{algorithm}

\subsection{Master level}\label{sec master_level}
At this level, the objective is to estimate the global row and column partition given sufficient statistics, local cluster sizes, and the prior. In the following, we detail these two steps: 
\subsubsection{Global row partition estimation} The global row membership $\mathbf{z}$ is estimated by clustering the local row  clusters. Instead of assigning the rows sequentially and individually to their row cluster, we assign the batch of rows that already share the same local row cluster to a global row cluster. Hence, the rows assigned to the same global row cluster will share the same global row membership. Since the workers operate asynchronously, the results are joined in a streaming way using the Reduce function without waiting for all workers to finish their tasks. 

Let $\mathcal{S}^{e_1}, S^{e_2}, K^{e_1}$ and $K^{e_2}$ be the sets of sufficient statistics and the number of local row clusters returned by two workers $e_1$ and $e_2$ respectively. The goal is to cluster the local row  clusters $\{\mathbf{x}_{1,\cdot}^{e_1},\cdots,\mathbf{x}_{K^{e_1},\cdot}^{e_1}\}$ and $\{\mathbf{x}_{1,\cdot}^{e_2},\cdots,\mathbf{x}_{K^{e_2},\cdot}^{e_2}\}$. To perform such clustering, we proceed as follows: we first set the initial cluster partition equal to the local partition inferred in cluster $e_1$. Then, for each $h\in\{1,\cdots, K^{e_2}\}$, we sample $z_h^{e_2}$, the membership of $\mathbf{x}_{h,\cdot}^{e_2}$ from $\mathrm{p}(z_h^{e_2}\mid \mathbf{z}^{e_2}_{-h}, X, G_0, \alpha) 
\propto$
\begin{subnumcases}{}
   $$n_{k} \mathrm{p}(\mathbf{x}^{e_2}_{h,\cdot} \mid z^{e_2}_h = k, \mathbf{X}_{k,\cdot}, G_0), $$ & existing row cluster $k$, 
   \label{eq m1bis}
   \\
   $$\alpha \mathrm{p}(\mathbf{x}^{e_2}_{h,\cdot} \mid G_0)$$ & new row cluster, \label{eq m1bis1}
\end{subnumcases}

where $n_{k}$ is the size of global row cluster $k$, $\mathbf{X}_{k,\cdot}$ the content of global row cluster $k$, and $\mathbf{z}^{e_2}_{-h} = \{{z}^{e_2}_{h'}|h'\neq h\}$. The joint posterior and the joint prior predictive distributions (Eq \ref{eq m1bis}, and Eq \ref{eq m1bis1} respectively) are computed analytically by only using sufficient statistics, i.e., without having access to the content of local and global clusters: 
$$
\mathrm{p}\left(\mathbf{x}_{h,\cdot}^{e} \mid G_0\right)=\pi^{-n_{h}^{e}\frac{d}{2}}\cdot \frac{\kappa_0^{d / 2}}
{\left(\kappa_{h}^{e}\right)^{d / 2}} \cdot \frac{\Gamma_d\left(\nu_{h}^{e} / 2\right)}{\Gamma_d\left(\nu_0 / 2\right)} 
\cdot \frac{\left|\Psi_0\right|^{\nu_0 / 2}}{\left|\Psi_{h}^{e}\right|^{\nu_{h}^{e} / 2}}
$$
where $|\cdot|$ is the determinant, $\Gamma$ denotes the gamma function, and the hyper-parameter $(\mu_h^e, \kappa_h^e, \Psi_h^e, \nu_h^e)$ are updated using the sufficient statistics:
$$
\begin{gathered}
\mu_{h}^{e}=\frac{\kappa_0 \mu_0+n_{h}^{e} T_{h}^{e}}{\kappa_{h}^{e}}, \quad \kappa_{h}^{e}=\kappa_0+n_{h}^ {e}, \quad \nu_{h}^{e}=\nu_0+n_{h}^{e}, \\
\Psi_{h}^{e}=\Psi_0+S_h^e+\frac{\kappa_0 n_{h}^{e}}{\kappa_{h}^{e}}\left(\mu_0-T_{h}^{e}\right)\left(\mu_0-T_{h}^{e}\right)^T,
\end{gathered}
$$
where $T_h^e = \frac{1}{p} \sum_{j = 1}^p T_{h,j}$ and $S_h^e = \frac{1}{p} \sum_{j = 1}^p S_{h,j}^e$. Moreover, we have  
$$
\mathrm{p}(\mathbf{x}^e_{h,\cdot} \mid z^e_h = k, \mathbf{X}_{k,\cdot}, G_0)=\pi^{\frac{-dn_{h}^{e}}{2}}\cdot \frac{\kappa_k^{d / 2}}{\left(\kappa_{h}^{e}\right)^{d / 2}} \cdot \frac{\Gamma_d\left(\nu_{h}^{e} / 2\right)}{\Gamma_d\left(\nu_k / 2\right)} \cdot \frac{\left|\Psi_k\right|^{\nu_k / 2}}{\left|\Psi_{h}^{e}\right|^{\nu_{h}^{e} / 2}}
$$
where the posterior distribution parameters $(\mu_k, \kappa_k, \Psi_k, \nu_k)$ associated to the global cluster $k$ are updated from the prior as follows:
$$
\begin{gathered}
\mu_{k}=\frac{\kappa_0 \mu_0+n_{k} T_k}{\kappa_{k}}, \quad \kappa_{k}=\kappa_0+n_{k}, \quad \nu_{k}=\nu_0+n_{k}, \\
\Psi_{k}=\Psi_0+S_k+\frac{\kappa_0 n_{k}}{\kappa_{k}}\left(\mu_0-T_k\right)\left(\mu_0-T_k\right)^T,
\end{gathered}
$$
with $T_k$ and $S_k$ the aggregated sufficient statistics when local clusters are assigned to the same global cluster. They are given by:  

\begin{eqnarray}
    &T_k = \frac{1}{n_k} \sum_{e,h|\,\mathbf{z_h^e=k}} n_h^e\cdot T_h^e, \label{aggssr1}\\ 
    &S_k = \sum_{e,h|\,\mathbf{z_h^e=k}} S_h^e + \sum_{e,h|\,\mathbf{z_h^e=k}}\left( n_e^h \cdot T_h^e\cdot {T_{h}^{e}}^T \right) - n_k \cdot T_k\cdot T_k^T.\label{aggssr2} 
\end{eqnarray}

\par \noindent This step consists of joining workers' local row  clusters in a streaming way. The recursive joining process stops when the global row  partition is estimated. If $K^{(e_1, e_2)}$ is the number of inferred global row clusters, then the process stops when $\sum_{k=1}^{K^{(e_1, e_2)}} n_k = n$. The procedure is detailed in the algorithm \ref{JOIN}.
\begin{algorithm}
\caption{Join workers results (Reduce Function)}\label{JOIN}
\begin{algorithmic}[1]
\State \textbf{Input}: 
$\mathcal{S}^{e_1}$, $\mathcal{S}^{e_2}$, $\alpha$ and prior $G_0$. 
\State Initialize global membership $\mathbf{z}$ according to $\mathbf{z}^{e_1}$.
\BState \textbf{For} each $h \in K^{e_2}$ \textbf{do}:
\BState \indent Sample $z_h^{e_2}$ according to Eq. \ref{eq m1bis} and  Eq. \ref{eq m1bis}.
\BState \indent Add $\mathbf{x}_h^{e_2}$ to its new global row  cluster.
\BState \indent Update the membership vector $\mathbf{z}$.
\BState \textbf{For} $k\leftarrow  1$ \textbf{to} $K^{(e_1, e_2)}$\textbf{do}:
\BState \indent Compute $S_k$ and $T_k$ according to Eq. \ref{aggssr1} and Eq \ref{aggssr2}.
\State \textbf{Output}: Updated row partition $\mathbf{z}$, aggregated sufficient statistics and clusters sizes.
\end{algorithmic}
\end{algorithm}
%
\subsubsection{Column memberships estimation}
Given the sufficient statistics $\mathcal{S}^1,\mathcal{S}^2,\cdots, \mathcal{S}^E$, the global row  partition $\mathbf{z}$, the prior $G_0$, and the concentration parameter $\beta$, the objective is to update the column partition $\mathbf{w}=(w_j)_p$, each $w_j$ is drawn according to $\textrm{p}(w_i|\mathbf{w}_{-j}, \mathbf{z}, X, G_0, \beta)\propto$ 
\begin{subnumcases}{}
   $$p_{k}\mathrm{p}(x_{\cdot,j} \mid \mathbf{z},\mathbf{w}_{-j}, X_{-j}, G_0, \beta), $$ & existing column cluster $l$, \label{eq m2bis}
   \\
   $$\beta \mathrm{p}(x_{\cdot,j} \mid \mathbf{z}, G_0)$$ & new column cluster, \label{eq m2bis1}
\end{subnumcases}
with $x_{.,j}$ the $j$-th column and $X_{-j}$ the dataset without column $j$. Similarly, the joint posterior predictive and the joint prior predictive distributions (Eq \ref{eq m2bis}, and Eq \ref{eq m2bis1} respectively) are computed analytically without having access to the columns, but only by using sufficient statistics. In fact, we have: 
$$\mathrm{p}\left(x_{\cdot, j} \mid \mathbf{z}, G_{0}\right)=\prod_{k=1}^K\mathrm{p} \left(\mathbf{x}_{k,j}|G_{0}\right)$$
with $K$ the global number of inferred row  clusters, and $\mathbf{x}_{k,j}$ the element of column $j$ that belong to the row  cluster $k$. We have: 
$$
\mathrm{p} \left(\mathbf{x}_{k,j}|G_{0}\right)=\pi^{-p_{k,j}\times\frac{d}{2}}\cdot \frac{\kappa_0^{d / 2}}{\kappa_{k,j}^{d / 2}} \cdot \frac{\Gamma_d(\nu_{k,j} / 2)}{\Gamma_d(\nu_0 / 2)} \cdot \frac{|\Psi_0|^{\nu_0 / 2}}{|\Psi_{k,j}|^{\nu_{k,j}/ 2}}
$$
with $p_{k,j}$ the cardinal of $\mathbf{x}_{k,j}$. The updated hyper-parameters are obtained with:
$$
\begin{gathered}
\mu_{k,j}=\frac{\kappa_0 \mu_0+p_{k,j} T_{k,j}}{\kappa_{k,j}}, \quad \kappa_{k,j}=\kappa_0+p_{k,j}, \quad \nu_{k,j}=\nu_0+p_{k,j}, \\
\Psi_{k,j}=\Psi_0+S_{k,j}+\frac{\kappa_0 p_{k,j}}{\kappa_{k,j}}\left(\mu_0-T_{k,j}\right)\left(\mu_0-T_{k,j}\right)^T.
\end{gathered}
$$
Moreover, the posterior predictive distribution is computed as follows:
$$
\mathrm{p}(x_{\cdot,j} \mid \mathbf{z},\mathbf{w}_{-j}, X_{-j}, G_0, \beta) = \prod_{k=1}^K\mathrm{p} \left(\mathbf{x}_{k,j}|G_{k,l}\right)
$$
with $G_{k,l}$ the posterior distribution associated with block $(k,l)$ (i.e., row  cluster $k$ and column cluster $l$). We have:
$$
\mathrm{p} \left(\mathbf{x}_{k,j}|G_{k,l}\right)=\pi^{-p_{k,j}\times\frac{d}{2}}\cdot \frac{\kappa_{k,l}^{d / 2}}{\kappa_{k,j}^{d / 2}} \cdot \frac{\Gamma_d(\nu_{k,j} / 2)}{\Gamma_d(\nu_{k,l} / 2)} \cdot \frac{|\Psi_{k,l}|^{\nu_{k,l} / 2}}{|\Psi_{k,j}|^{\nu_{k,j}/ 2}}
$$
with $(\mu_{k,l},\kappa_{k,l},\Psi_{k,l},\nu_{k,l})$ the block posterior distribution parameters given by:
$$
\begin{gathered}
\mu_{k,l}=\frac{\kappa_0 \mu_0+p_{k,l} T_{k,l}}{\kappa_{k,l}}, \quad \kappa_{k,l}=\kappa_0+p_{k,l}, \quad \nu_{k,l}=\nu_0+p_{k,l}, \\
\Psi_{k,l}=\Psi_0+S_{k,l}+\frac{\kappa_0 p_{k,l}}{\kappa_{k,l}}\left(\mu_0-T_{k,l}\right)\left(\mu_0-T_{k,l}\right)^T.
\end{gathered}
$$
\par \noindent With $T_{k,l}$ and $S_{k,l}$, the aggregated sufficient statistics obtained when local clusters are assigned to the same global block $(k,l)$, and they are computed as follows: 
$$
\begin{gathered}\label{ST_K_J}
    T_{k,l} = \frac{1}{p_{k,l}} \sum_{e,h|\,\mathbf{z_h^e=k},\mathbf{w=l}} p_{h,l}^e\cdot T_{h,l}^e \label{eq T_K_L}\\
    S_{k,l} = \sum_{e,h|\,\mathbf{z_h^e=k},\mathbf{w=l}} S_{h,l}^e + \sum_{e,h|\,\mathbf{z_h^e=k},\mathbf{w=l}}\left( p_{h,l}^e \cdot T_{h,l}^e\cdot {T_{h,l}^{e}}^T \right) - p_{k,l} \cdot T_{k,l}\cdot T_{k,l}^T \label{eq S_K_L}
\end{gathered}
$$
where $p_{k,l}$ is the number of cells in the global cluster $(k,l)$. The column partition update is detailed in algorithm \ref{AGREGATOR}.
\begin{algorithm}
\caption{Column clustering}\label{AGREGATOR}
\begin{algorithmic}[1]
\State \textbf{Input}: Sufficient statistics, row partition, $\beta$ and prior $G_0$. 
\BState \textbf{For} $j\leftarrow  1$ \textbf{to} $p$ \textbf{do}:
\BState\indent Remove $x_{.,j}$ from its column cluster.
\BState\indent Sample $w_{j}$ according to Eq \ref{eq m2bis}, and Eq \ref{eq m2bis1}.
\BState\indent Add $x_{.,j}$ to its new column cluster.
\State \textbf{Output}: Column-partition $\mathbf{w}$.
\end{algorithmic}
\end{algorithm}

\section{Experiments}
To evaluate our approach, we conducted several experiments. Firstly, we compare our distributed algorithm with other state-of-the-art co-clustering and clustering algorithms in terms of row clustering performance on synthetic and real-world datasets. Secondly, we compare the execution time and clustering performance of our distributed algorithm DisNPLBM and the centralized NPLBM \cite{goffinet2021coclustering} on synthetic datasets with different row sizes. Lastly, we investigate the scalability of DisNPLBM by increasing the number of nodes while keeping the number of rows fixed. The clustering performance is evaluated using the clustering metrics Adjusted Rand Index (ARI) \cite{hubert1985comparing} and Normalized Mutual Information (NMI)\cite{nmi}.

\subsection{Experiment settings}
In the following experiments, we use an uninformative prior $\textrm{NIW}$ as in \cite{goffinet2021coclustering}. Therefore, we set the $\textrm{NIW}$ hyper-parameters as follows: $\mu_0$, and the matrix precision $\Psi_0$ are respectively set to be empirical mean vector and covariance matrix of all data. $\kappa_0$ and $\nu_0$ are set to their lowest values, which are $1$ and $d+1$, respectively, where $d$ is the dimension of the observation space. The initial partition consists of a single cluster, and the algorithms run for $100$ iterations.

The distributed algorithm is executed on the Neowise machine (1 CPU AMD EPYC 7642, 48 cores/CPU) and Gros machines (1 CPU Intel Xeon Gold 5220, 18 cores/CPU), both hosted by Grid5000\footnote{\url{https://www.grid5000.fr/}}. For enhanced portability and deployment flexibility, DisNPLBM is containerized using the Docker image bitnami/spark 3.3.0. We employ Kubernetes for orchestrating Docker images and deploy the Kubernetes cluster on Grid5000 using Terraform\footnote{\url{https://www.terraform.io/}}.

\subsection{Clustering performance}
We first evaluate the row clustering performance of our algorithm on both synthetic and real-world datasets; we compare its results with two co-clustering algorithms, NPLBM \cite{goffinet2021coclustering} and LBM \cite{article}, and two clustering algorithms, K-means and  Gaussian mixture model (GMM). We applied the algorithms to $4$ datasets: \textit{Synthetic} dataset of size $150 \times 150 \time 1$, generated from $10 \times 3$ Gaussian components. \textit{Wine} dataset \cite{misc_wine_109} represents a chemical analysis of three types of wines grown in the same region. The dataset consists of $178$ observations, $12$ features, and $3$ clusters. We also apply the algorithms to two bioinformatics datasets \textit{Chowdary} ($104$ samples, $182$ genes, and $2$ clusters) \cite{Chowdary2006} and \textit{Nutt} ($22$ samples, $1152$ genes, and $2$ clusters) \cite{Nutt2003}. Each sample's gene expression level is measured using the Affymetrix technology leading to strictly positive data ranging from $0$ to $16000$. we apply the Box-Cox transformation \cite{box1964analysis}, to make the data  Gaussian-like. Since the number of Genes is much greater than the number of samples we distribute the columns across the workers to achieve scalability, this is legitimate since the row and column clustering are symmetric in our case. 

\begin{figure}[t!]
    \centering
    \includegraphics[scale=0.30]{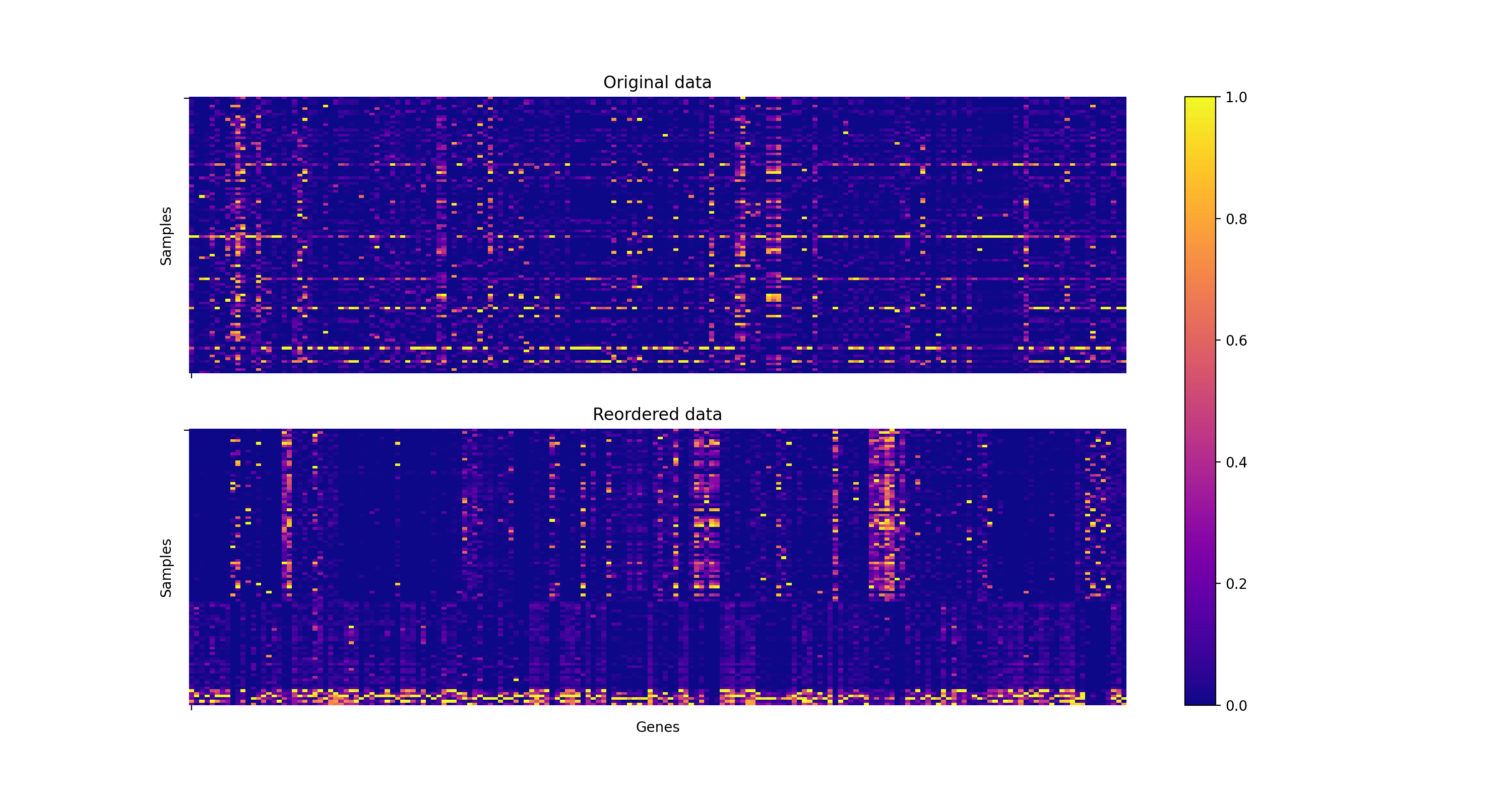}
    \caption{Heatmaps of Chowdary data. The first row represents the original data. The second row represents the reordered data after DisNPLBM.}
    \label{fig:cc}
\end{figure}

Table \ref{clustresults} presents the mean and standard deviation of ARI and NMI across $10$ launches for each method on each dataset. Our method outperformed other approaches in the Bioinformatics datasets. Additionally, it has estimated the true clustering structure in the synthetic dataset. While NPLBM and LBM slightly outperform our method on the \textit{Wine} dataset, our approach still yields satisfying results, surpassing traditional methods like GMM and K-means. It's crucial to note that this experiment focuses on comparing clustering performance, without considering execution times due to different inference algorithms. Figure \ref{fig:cc} illustrates the Heatmaps of Chowdary data before DisNPLBM and reordered data after DisNPLBM. In the recorded data, there is a visible checkerboard pattern distinguishing co-clusters. Co-clustering simultaneously clusters samples and genes, revealing groups of highly correlated genes with distinct correlation structures among different sets of individuals, such as between disease and healthy individuals or different types of disease. This may allow to identify which genes are responsible for some diseases. 

\begin{table}[t!]
  \centering
  \begin{tabular}{llccccc}
    \toprule
    {Dataset}                  &       & {DisNPLBM} & {NPLBM} & LBM & GMM & K-means\\
    \midrule
       \multirow{2}{*}{Synthetic}  & ARI & $\mathbf{1.00 \pm 0.00}$ & $\mathbf{1.00 \pm 0.00}$  & $0.42 \pm 0.03$ & $0.38 \pm 0.05$ & $ 0.39 \pm 0.01 $ \\
                               & NMI & $\mathbf{1.00 \pm 0.00}$ & $\mathbf{1.00 \pm 0.00}$ & $0.78 \pm 0.02$ & $0.70 \pm 0.02$ & $0.71 \pm 0.01$\\
    \midrule
       \multirow{2}{*}{Wine}   & ARI & $0.52 \pm 0.03 $ & $\mathbf{0.56 \pm 0.04}$ &  $0.56 \pm 0.07$ & $0.51 \pm 0.07$ & $0.50 \pm 0.04$ \\
                               & NMI & $0.59 \pm 0.02$ & $\mathbf{0.65 \pm 0.03}$ & $0.64 \pm 0.03$ & $0.64 \pm 0.03$ & $0.64 \pm 0.02$ \\      
    \midrule
       \multirow{2}{*}{Chowdary}   & ARI & $ \mathbf{0.78 \pm 0.01} $ & $ 0.07\pm 0.01$ &  $ 0.65\pm 0.00 $ & $0.74 \pm 0.01 $ & $ 0.75 \pm 0.01 $ \\
                               & NMI & $ \mathbf{0.68 \pm 0.02}$ & $ 0.11 \pm 0.01 $ & $ 0.58\pm 0.01$ & $ 0.63 \pm 0.01$ & $ 0.64 \pm 0.01 $ \\      
  \midrule
       \multirow{2}{*}{Nutt}   & ARI & $\mathbf{0.58 \pm 0.01}$ & $ 0.56 \pm 0.02 $ &  $ 0.54 \pm 0.04$ & $0.08 \pm 0.00$ & $ 0.11\pm 0.04$ \\
                               & NMI & $\mathbf{0.74 \pm 0.01}$ & $0.74 \pm 0.01$ &  $ 0.68\pm 0.02$ & $0.28 \pm 0.02$ & $ 0.30\pm 0.00 $ \\      

 \bottomrule    
 \end{tabular}
\caption{The mean and the standard deviation of ARI and NMI over $10$ runs on different datasets. The best result within each row is marked as bold.}
\label{clustresults}
\end{table}
\subsection{Comparison of the distributed and centralized approaches}
We compare the execution times and clustering performance of the distributed and the centralized 
NPLBM. We execute both algorithms on synthetic datasets of sizes $n\times p\times d$, where $n\in\{20\mathrm{K},\cdots, 
100\mathrm{K}\}$, $p=90$ and $d=1$ generated from $K\times L$ Gaussian components, with $K=10$ and $L=3$ (i.e., $K=10$ 
row clusters and $L=3$ column clusters). We stop at $n=100$K because the centralized version is too slow; running over $100$K observations would take too much time. The distributed algorithm is executed on the Neowise machine in local mode using $24$ cores. The centralized algorithm is executed on the same machine using one core. 
\begin{table}[H]
    \centering
    \begin{tabular}{lcccccccc}
     \toprule
      \multirow{2}{*}{$n$} &
      \multicolumn{2}{c}{ARI} &
      \multicolumn{2}{c}{NMI} &
      \multicolumn{2}{c}{$\hat{K}\times\hat{L}$} &
      \multicolumn{2}{c}{Running time (s)} \\
      & {Dis.} & {Cen.} & {Dis.} & {Cen.} & {Dis.} & {Cen.} & {Dis.} & {Cen.}\\
      \midrule
      20K  & 1.0 & 1.0 & 1.0 & 1.0 & 30 & 30 & 400.21  & 2265.69 \\
      40K  & 1.0 & 1.0 & 1.0 & 1.0 & 30 & 30 & 693.02  & 6452.78 \\
      60K  & 1.0 & 1.0 & 1.0 & 1.0 & 30 & 30 & 1122.80 & 10511.01 \\
      80K  & 1.0 & 1.0 & 1.0 & 1.0 & 30 & 30 & 1373.04 & 19965.01 \\
      100K & 1.0 & 1.0 & 1.0 & 1.0 & 30 & 30 & 1572.90 & 41897.12\\    
     \bottomrule
    \end{tabular}
 \caption{ARI, NMI, number of inferred block clusters ($\hat{K}\times\hat{L}$), and the running time in seconds achieved by the distributed (Dis.) and centralized (Cen.) algorithms.}
 \label{table:DisVsCent}
\end{table}
Table \ref{table:DisVsCent} reports the clustering metrics ARI, NMI, number of inferred block clusters ($\hat{K}\times\hat{L}$), and the running times obtained by the centralized and distributed inference algorithms on datasets with different row sizes. The results show that our approach considerably reduces the execution time. For example, it is reduced by a factor of $26$ for a dataset with $100$K rows. On the other hand, we remark that both the cen distributed and centralized methods performed very well in terms of clustering with values of 1 indicating perfect clustering. Moreover, both methods inferred the true number of clusters. Overall, the distributed approach runs much faster than the centralized method without compromising the clustering performance which makes it more efficient in terms of computational time.

\subsection{Distributed Algorithm Scalability}
We now investigate the scalability of our approach by increasing the number of cores up to 64 in a distributed computing environment. We employ a dataset with $n=500$K rows, $p=20$ columns, and $d=1$ (representing the observation space dimension). The dataset is generated from $K\times L$ Gaussian components, where $K=10$ is the number of row clusters and $L=3$ is the number of column clusters. To conduct this evaluation, we deploy a Kubernetes cluster using up to 6 Gros Machines.
\begin{table*}
\centering
\begin{tabular}{cllcc}
\toprule
{Cores} & {ARI} & {NMI} & {$\hat{K}\times \hat{L}$} & {Running time (s)}  \\  
\midrule
{2}  & 1.0  & 1.0  & 30 & {88943.45} \\ 
{4}  & 1.0  & 1.0  & 30 & {27964.73} \\ 
{8}  & 0.99 & 0.99 & 30 & {16202.15} \\
{32} & 0.98 & 0.99 & 33 & {2715.80}  \\
{64} & 0.98 & 0.99 & 33 & {1861.85}  \\
 \bottomrule
  \end{tabular}
      \caption{ARI, NMI, number of inferred block clusters ($\hat{K}\times\hat{L}$), and the running time in seconds achieved by the distributed approach when distributing on different number of cores.}
      \label{tabl:Cores}
\end{table*}
Table \ref{tabl:Cores} presents clustering metrics ARI and NMI, the number of inferred block clusters, and running time as the number of cores increases. The running time significantly decreases with an increasing number of cores, with the execution time reduced by a factor of 48 when using 64 cores compared to two cores. This demonstrates the efficient scalability of our algorithm with the number of workers. It's worth noting a slight overestimation of the number of clusters with more cores. Additionally, there is a slight decrease in ARI and NMI scores. Nevertheless, our approach still achieves very high clustering metrics and accurately estimates the number of clusters.

\section{Conclusion}

This article presents a novel distributed MCMC inference for NPLBM. NPLBM has the advantage of estimating the number of row and column clusters. However, the inference process becomes too slow when dealing with large datasets. Our proposed method achieves high scalability without compromising the clustering performance. Our future research will explore the potential extension of this method to the multiple Coclustering model.

\section{Acknowledgements}
This work has been supported by the Paris Île-de-France Région in the framework of DIM AI4IDF. I thank Grid5000 for providing the essential computational resources and the start-up
 HephIA for the invaluable exchange on scalable algorithms.

\bibliographystyle{unsrt}  
\bibliography{bibliography}  
\end{document}